\documentclass[conference]{IEEEtran}
\IEEEoverridecommandlockouts
\usepackage{graphicx,booktabs,amsmath,amssymb,url,color,flushend}
\usepackage[linesnumbered,ruled]{algorithm2e}
\usepackage[misc]{ifsym}

\usepackage{subfig}
\usepackage{floatrow}    

\usepackage{bm}
\usepackage{multirow}
\usepackage[noend]{algpseudocode}               
\algnewcommand\INPUT{\item[\textbf{Input:}]}
\algnewcommand\OUTPUT{\item[\textbf{Output:}]}
\def\BibTeX{{\rm B\kern-.05em{\sc i\kern-.025em b}\kern-.08em
		T\kern-.1667em\lower.7ex\hbox{E}\kern-.125emX}}
\begin{document}
	\title{Classification of Noncoding RNA Elements Using Deep Convolutional Neural Networks
	}
	
	\author{\IEEEauthorblockN{Brian McClannahan, 	
	Krushi Patel, Usman Sajid, Cuncong Zhong, Guanghui Wang}
		\IEEEauthorblockA{\textit{Department of Electrical Engineering and Computer Science}\\
			\textit{The University of Kansas,
				Lawrence, KS, USA 66045} \\
			\{b523m844, usajid, cczhong, ghwang\}@ku.edu}
	}

\maketitle

\begin{abstract}
The paper proposes to employ deep convolutional neural networks (CNNs) to classify noncoding RNA (ncRNA) sequences. To this end, we first propose an efficient approach to convert the RNA sequences into images characterizing their base-pairing probability. As a result, classifying RNA sequences is converted to an image classification problem that can be efficiently solved by available CNN-based classification models. The paper also considers the folding potential of the ncRNAs in addition to their primary sequence. Based on the proposed approach, a benchmark image classification dataset is generated from the RFAM database of ncRNA sequences. In addition, three classical CNN models have been implemented and compared to demonstrate the superior performance and efficiency of the proposed approach. Extensive experimental results show the great potential of using deep learning approaches for RNA classification.
\end{abstract}


\section{Introduction}
\label{sec:introduction}

In 2007, with the development of high-throughput sequencing, a large amount of new RNA were discovered \cite{$graphclust$}. Many of these RNA were found not to be involved in protein creation and are thus called noncoding RNA (ncRNA). Noncoding RNA plays many important roles in cell processes \cite{$review$}. For example, microRNA (miRNA) performs posttransciptional regulation of gene expression, while long noncoding RNAs can regulate epigenetic modification and gene expression \cite{$review$}. The classification of ncRNAs seeds to categorize ncRNA elements into families based on their sequence and structure to facilitate their functional annotation and prediction \cite{$review$}.

The molecular function of ncRNA is implemented through both of its sequence and structure \cite{$graphclust$}. Classification of ncRNAs purely using their sequences is insufficient, as ncRNAs with conserved secondary structures may share low sequence identitiy due to the presence of covariant mutations. Therefore, modern methods for ncRNA clustering use both the primary sequence and secondary structure features \cite{$dotaligner$}. However, the consideration of the secondary structure increases the time complexity for pairwise ncRNA comparison, from $O(l^2)$ with pure sequence to $O(l^4)$ with both sequence and secondary structure. The high time complexity thus makes clustering of large amount of ncRNA elements infeasible. To address this issue, we turn to machine learning approaches, whose classification phase is very efficient when the model is trained. 

\begin{figure}[tp]
	\captionsetup[subfigure]{labelformat=empty}
	\subfloat[]{\includegraphics[width=10em]{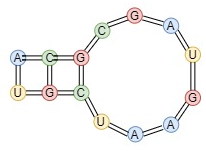}}\\[-8pt]
	\subfloat[]{\includegraphics[width=10em]{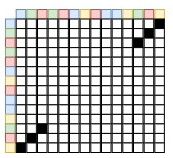}}
	\caption{Top: an example of interior loop of an RNA sequence. Bottom: Corresponding dot-plot matrix of the sequence.}
\end{figure}

In this paper, we propose a new approach to general RNA family classification using image processing techniques. For a given ncRNA element, we computed its base-pairing probability matrix (BPPM) using RNAfold from the Vienna Package. We then convert the BPPM into an gray-scale image, using the intensity of each pixel to represent the base-pairing probability of the corresponding bases. We then apply three different deep CNN algorithms (VGGNet-19, ResNet-50, and ResNet-101) to classify these images. We tested our approach using the RFAM database \cite{$rfam$}, and showed an 85\% classification accuracy.

The main contribution of this paper include:
\begin{enumerate}
    \item We propose, for the first time, to convert the problem of RNA sequence classification into an image classification problem.
    \item We propose an efficient approach to convert two RNA sequences into an image and generate an image dataset for RNA sequences from the same and different families.  
    \item We implement three classical deep learning-based classification models and compare their performance in RNA classification. The results demonstrate the feasibility of the advantages of the proposed approach.
\end{enumerate}

The rest of this paper is organized as below. Section II reviews some related work on RNA analysis and image classification. Section III presents the approach and process to generate the image dataset from RNAs. The implemented CNN models and experimental results are presented in Sections IV and V, respectively. Finally, the paper is concluded in Section VI.

\section{Related Work}
\label{sec:related_work}
\noindent\textbf{Analyzing RNA by secondary structure:} There are several methods that are used for classifying RNA by its secondary structure. Some utilize context-free grammars, often paired with hidden Markov models, such as QRNA\cite{$qrna$}. Many will use secondary structure comparisons based on minimum free energy model like RNAfold \cite{$rnafold$} and Mfold \cite{$mfold$}, by predicting the conserved structure of RNA based on its thermodynamic stability \cite{$rnaz$}, or with a folding algorithm like Sankoff-style simultaneous alignment and folding algorithm \cite{$sankoff$} used by CARNA \cite{$carna$}. Several approaches comparing graph representations of the secondary structure such as GraphClust \cite{$graphclust$} and \cite{$graph_2004$}. Lastly, by combining sequence alignment techniques with a partition function, like the McCaskill Partition function \cite{$mccaskill$}, to compare the primary and secondary structures at the same time, such as LocaRNA-P \cite{$locarna$} and DotAligner \cite{$dotaligner$}. Before LocaRNA, classifiers struggled to accurately analyze ncRNA. DotAligner uses both the primary sequence and secondary structure to discover sequence motifs in lncRNAs \cite{$dotaligner$}. DotAligner first aligns the sequence, then applies a partition function to calculate the most likely secondary structures, and then maps the alignment to the predicted secondary structure to find sequence motifs to make predictions on RNA clusters, catching classifications that just a sequence comparison approach would miss. In this study, we propose to solve this from a new perspective of image classification using deep neural networks. 

	\qquad

\begin{figure}[htp]
	\captionsetup[subfigure]{labelformat=empty}
	\subfloat[a]{\includegraphics[width=10em]{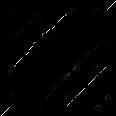}}
	\qquad
	\subfloat[b]{\includegraphics[width=10em]{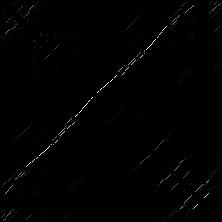}}\\
	\subfloat[c]{\includegraphics[width=10em]{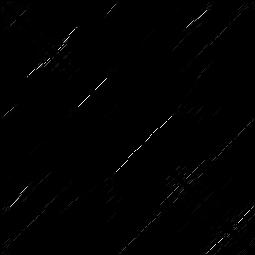}}
	\qquad
	\subfloat[d]{\includegraphics[width=10em]{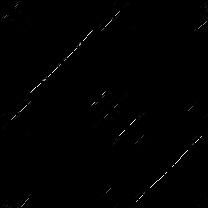}}
	\caption{Four dot-plot examples of RNA secondary structure  RFAM families a) 5S rRNA, b) C0719, c) snR63, d) S\_pombe\_snR42}
	\label{ref_label_overall}
\end{figure}

\noindent\textbf{Image classification and CNNs} Recently, researchers have applied CNNs to many different fields of applications, including image classification \cite{cen2019boosting}, translation \cite{xu2019adversarially}, object detection \cite{ma2020mdfn}, depth estimation \cite{he2018spindle}, crowd counting \cite{sajid2020zoomcount}, and medical image analysis \cite{patel2020comparative}. AlexNet applied CNNs to image classification and won the ILSVRC-2012 competition with a top-5 error rate of 15.3\% \cite{$alexnet$}. GoogLeNet introduced the inception module which employs $3 \times 3$ and $5 \times 5$ convolution masks followed be $1 \times 1$ convolution masks for dimensionality reduction and reduce the total number of parameters required to be trained by the network \cite{$googlenet$}. VGGNet showed how larger convolutional filters could be implemented more efficiently as stacked $3 \times 3$ convolutional filters, and achieved the first place in localization and the second in classification at ImageNet Challenge 2014 with a top-5 error rate of 7.3\% \cite{$vggnet$}. ResNet introduced the residual mappings to CNNs which allowed training much deeper networks without overfitting, achieving the first in ILSVRC 2015 classification task with a top-5 error rate of 6.71\% \cite{$resnet$}. After ResNet created significantly deeper networks, wide residual networks showed that similar results could be achieved by greatly increasing the number of parameters in a shallow network and utilizing dropout layers \cite{$wrn$}. Because of their success in image classification and popularity in deep learning, we chose to use both VGGNet and ResNet deep learning models for RNA classification.

Very few study has been performed for RNA classification. CNNclust \cite{$cnnclust$} utilized one-dimensional CNNs for ncRNA sequence motif discovery. Similar to DotAligner, CNNclust used both the sequence and the secondary structure to more accurately find sequence motifs and then use those to make accurate classifications. However, CNNclust utilized one-dimensional convolutions over the primary sequence and base-pair probability as opposed to using partition functions like LocaRNA and DotAligner. Instead of applying one-dimensional convolutions, we convert the secondary structure into an image so we can take advantage of image classification CNNs. These CNNs employ stacked layers of convolutional filters to extract the feature maps, then analyze these features with a fully-connected network to classify the image.

\section{Dataset Generation}
\label{sec:dataset}
The dot-plot matrix is generated by letting any cell $(i,j)$ represent the probability that a bond exists between the $i$th and $j$th nucleic acid in the RNA secondary structure (excluding bonds amongst neighboring nucleic acids). Generating a dot-plot matrix from the secondary structure gives a matrix that is symmetric along the diagonal, as shown in Fig. 1. The dot-plot matrix is then converted into an image by treating each cell as a pixel where the probability of a bond forming is treated as an intensity between 0 and 1, as illustrated in Fig. 2. This process creates a grey-scale, symmetric image representation of the secondary structure of any given RNA. 


One important research in RNA classification is to determine two RAN sequences are from the ``same family" or ``different family". To solve this problem, we need to build a network with two inputs for the two RNA sequences, which is practically hard to implement and train the network since there are numerous RNA sequences. In this study, we propose to convert the problem into an image classification problem. As illustrated in Fig. 3, we make use of the symmetric property of the dot-plot of each sequence and generate one single image from the two sequences, with the top-right half from sequence 1 and the lower-left half from sequence 2. Some generated examples are shown in Fig. 4. In this way, the problem is converted to determining if the generated belongs to the category of ``same family" or ``different family". Therefore, we can make use of all available image classification models from RNA classification. We can also see from Fig. 4 that the images from the same family have better symmetric property than those from different families, however, the differences are very small. The models are trained to distinguish these small changes.

\begin{figure}[htp]
	\captionsetup[subfigure]{labelformat=empty}
	\subfloat[]{\includegraphics[width=17em]{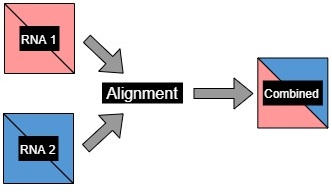}}
	\caption{Combine two dot-plot matrices into one image for classification.}
\end{figure}
\begin{figure}[htp]
	\captionsetup[subfigure]{labelformat=empty}
	\subfloat[a]{\includegraphics[width=10em]{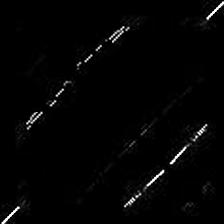}}
	\qquad
	\subfloat[b]{\includegraphics[width=10em]{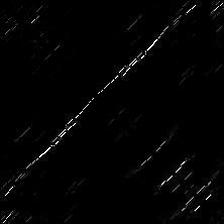}}\\
	\subfloat[c]{\includegraphics[width=10em]{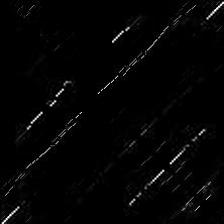}}
	\qquad
	\subfloat[d]{\includegraphics[width=10em]{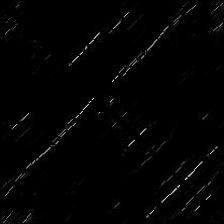}}
	\caption{Sample images from two RNA sequences. (a) and (b) are examples of two RNAs from the same family; (c) and (d) are example of RNA from different families.}
	\label{ref_label_overall}
\end{figure}

To train and evaluate each deep learning model, a large dataset is required. In this paper, we select all RNA families from the RFAM database that had a sequence length between 200 and 260 for the convenience of study. We resize the dot-plots of all sequences to $224 \times 224$ so as to combine each pair of them into an image. There were 168 families in this range. If we create the dataset exhaustively to cover all possible RNA combinations, the resulting dataset will have over 17 million images, which is not necessary for model training as there are a lot of redundant information in the dataset. The largest family size in this set was 712 while many of the families only had 2. In order to further reduce the dataset and balance the varying family sizes, each family larger than 30 RNA was truncated to 30. The 30 RNA picked from each family were chosen randomly. This reduced the total number of images to slightly greater than 2.5 million. The 168 families were split into the train/val/test set at a ratio of 70:10:20, resulting in a family split of 121/19/28, as shown in Table 1.

\begin{table}[ht]
	\begin{tabular}{ |l|l|l|l| } 
		\hline
		Set & \# of Families & Class & Image Count \\ 
		\hline
		\multirow{2}{*}{Train} 
		& \multirow{2}{*}{121} 
		& Same & 30,522 \\ 
		& & Different & 290,400\\ 
		\hline 
		\multirow{2}{*}{Val} 
		& \multirow{2}{*}{19}
		& Same & 5,374 \\
		& & Different & 68,882 \\
		\hline
		\multirow{2}{*}{Test} 
		& \multirow{2}{*}{28}
		& Same & 9,520 \\
		& & Different & 178,402 \\
		\hline
	\end{tabular}
	\caption{Final image count of the dataset before the iterative random image selection algorithm was applied.}
\end{table}

In table I, we have generated the images using all possible combinations of the RNA sequences except for the training data of different families. With 121 families in the training set, we can generate over 2.4 million training images for sequences from different families is we use all possible combinations of RNAs. However, in our experiments, we found that the models converged quickly without using all the training data. To increase the representativeness and reduce the size of the training data, we adopt the following approach to create the training images of different families. First, we randomly select an RNA from each family. Then, we randomly pick an RNA from every other family and combine them together to generate a training image of different families. We repeat this process 20 times and generate a total of 290,400 different family images of the final training set. 
\section{Classification Models}
\label{sec:models}
Convolutional neural networks (CNN) contain two parts, a feature extractor and a fully-connected network. The feature extractor is composed of convolutional layers mixed with activation and pooling layers. The convolutional layers, serving as feature extractors, are each followed by an activation layer, typically ReLU activation, to allow for the extraction of nonlinear features \cite{li20202}. The pooling layers, which typically come at the end of the convolutional layers, or at the end of a block of convolutional layers depending on the model design, reduce the spatial resolution of the feature maps, making the model more robust to input distortions \cite{$convolutional_review$}. The fully-connected network is a densely connected network that interprets the feature representations extracted by the feature network.

In the experiment, two classical CNN-based classification models were implemented: VGGNet \cite{$vggnet$} and ResNet \cite{$resnet$}. VGGNet was showed how larger convolution filters could be implemented more efficiently using stacked 3x3 convolutions. It achieved first in the ImageNet Challenge 2014 localization task and second in the classification task with VGGNet-19 scoring 7.5\% top-5 error rate. We selected the largest VGGNet model, VGGNet-19, a network with 16 convolution layers and 3 fully-connected layers, because it has the best performance of all the VGGNet networks.

ResNet \cite{$resnet$} is an improvement on the VGGNet \cite{$vggnet$} based scheme. ResNet introduced the concept of residual mapping in the deep learning field. Convolutional Neural networks attempt to approximate a function, but as the network gets deeper they can become increasingly hard to train. By inserting these residual maps throughout the network, ResNet was able to achieve significantly deeper networks than its predecessors. Two ResNet models were picked, ResNet-50 ResNet-101. ResNet achieved first in the ImageNet 2015 classification task with ResNet-50 and ResNet-101 scoring 5.25\% and 4.60\% top-5 error rate respectively. ResNet also achieved first in the ImageNet 2015 detection and localization tasks.
\section{Experimental analysis}
\label{sec:experiment}

In the experiments, we compared the performance of three models: VGGNet-19, ResNet-50, and ResNet-101. All models employ the SGD optimizer with a momentum factor of 0.9. The initial learning rates of ResNet and VGGNet-19 are set at 0.01 and 0.005, respectively, and the learning rates are reduced by a factor of 0.25 every 50 iterations. The batch size is set at 320. At each iteration, the model is trained on 320 images, selected randomly from the training dataset. Each training iteration took 18 seconds (all training was done on an nvidia k40 GPU). The models are validated every 50th iteration on the entire validation set. Each validation epoch took 15 minutes. Training for each model took approximately six hours. Both ResNet models were trained for 600 iterations, and VGGNet-19 was trained for 700 iterations. After training, we choose the model with the best performance on the validation set. Then, we evaluate their performance using the average class accuracy on the test set.

As shown in Table I, the two classes of the dataset are highly imbalanced. The size of different-family images is much larger than that of the same family. In order to reduce the influence of the class imbalance in the training set, we choose different class ratios at the training stage. We choose three Different-same ratios: 1:1, 2:1, and 4:1, as shown in Table II. At the ratio of 1:1, for the batch of 320 training images at each iteration, we randomly select the same number of images from the two classes. However, at the ratio of 4:1, we randomly selected 256 and 64 images from the classes of different-family and same-family respectively so as to ensure more different-family images are involved in the training process. The different ratios only apply to the training, while for validation and test, we always use the same ratio in our experiments.


\begin{table}[]
	\begin{tabular}{|l|l|c|c|c|}
		\hline
		\multicolumn{5}{|c|}{Accuracy}                                                                                                                      \\ \hline
		\multicolumn{2}{|l|}{Model}                        & \multicolumn{1}{l|}{VGG-19} & \multicolumn{1}{l|}{ResNet-50} & \multicolumn{1}{l|}{ResNet-101} \\ \hline
		\multicolumn{5}{|c|}{Diff-Same Ratio: 1-1}                                                                                                          \\ \hline
		\multicolumn{2}{|l|}{Train}                        & 92.3\%                        & 94.7\%                           & 96.9\%                            \\ \hline
		\multicolumn{2}{|l|}{Val}                          & 78.2\%                        & 78.8\%                           & 80.0\%                            \\ \hline
		\multicolumn{1}{|c|}{\multirow{3}{*}{Test}} & Diff & 80.0\%                        & 83.0\%                           & 84.0\%                            \\ \cline{2-5} 
		\multicolumn{1}{|c|}{}                      & Same & 85.0\%                        & 82.0\%                           & 83.5\%                            \\ \cline{2-5} 
		\multicolumn{1}{|c|}{}                      & Avg  & 82.5\%                   & 82.5\%              & \textbf{83.5\%}                 \\ \hline
		\multicolumn{5}{|c|}{Diff-Same Ratio: 2-1}                                                                                                          \\ \hline
		\multicolumn{2}{|l|}{Train}                        & 91.9\%                        & 92.8\%                           & 96.6\%                            \\ \hline
		\multicolumn{2}{|l|}{Val}                          & 79.2\%                        & 81.5\%                           & 80.8\%                            \\ \hline
		\multirow{3}{*}{Test}                       & Diff & 84.0\%                        & 86.0\%                           & 86.0\%                            \\ \cline{2-5} 
		& Same & 84.0\%                        & 84.0\%                           & 83.0\%                            \\ \cline{2-5} 
		& Avg  & 84.0\%                      & \textbf{85.0}\%                           & 84.5\%                    \\ \hline
		\multicolumn{5}{|c|}{Diff-Same Ratio: 4-1}                                                                                                          \\ \hline
		\multicolumn{2}{|l|}{Train}                        & 90.0\%                        & 95.6\%                           & 95.6\%                            \\ \hline
		\multicolumn{2}{|l|}{Val}                          & 77.5\%                        & 80.5\%                           & 80.0\%                            \\ \hline
		\multirow{3}{*}{Test}                       & Diff & 87.0\%                        & 89.0\%                           & 90.0\%                            \\ \cline{2-5} 
		& Same & 78.0\%                        & 81.0\%                           & 80.0\%                            \\ \cline{2-5} 
		& Avg  & 82.5\%                        & \textbf{85.0\%}                & \textbf{85.0\%}                 \\ \hline
	\end{tabular}
	\caption{Comparative results of different models trained on different dataset ratios.}
\end{table}

The comparative results can be found in Table II. Both ResNet models achieved a similar average class accuracy, 84.5\% and 85\% for ResNet-50 and ResNet-101 respectively. The VGGNet-19 model performed slightly worse with a top accuracy of 83\%. As evidenced in the results, a higher Diff:Same ratio resulted in the model improving its accuracy on the different classes while decreasing its performance on the same class. It is difficult to directly compare our network to other approaches because of the differences in datasets. The closest approach in the literature is the most recent CNNclust \cite{$cnnclust$}, which employs one-dimensional convolutions on the primary and secondary sequences to look for common patterns in both. CNNclust only achieved 75.2\% clustering accuracy when tested on ncRNA families not from the training set \cite{$cnnclust$}. We are able to achieve better performance by utilizing two-dimensional convolutions on just the secondary structure. Once the models are trained, the models can process around 80 RNA comparisons per second. However, traditional approaches are very time-consuming since they have to perform pairwise alignment on their base-pairing probability matrices, which has a time complexity of $O(l^4)$ while $l$ is the lengths of the ncRNA sequences.

This study demonstrates the potential of applying CNN-based image classification models to RNA classification. In order to further improve these results, a few measures can be taken. One approach is to increase the dataset to include more RNA families by removing the limitations on sequence length. The distortion caused by increasing the disparity in sequence length may lower the accuracy. However, RNA sequence length amongst families does not vary much so it is more likely that the CNN's feature network recognizes this distortion as a strong indicator of different classes. Another possible measure is applying a clustering algorithm to the classification results. These clusters can be determined by groups of RNA with a high percentage of same-family classifications amongst themselves. Clustering would also give more meaningful data to tune the hyper-parameters. We can also adopt a better approach to handle the imbalance issue of the dataset. As shown in Table II, when we choose 4-1 Diff:Same ratio, the test accuracy for different families is close to 90\%, which means there are still rooms to further increase the performance of the CNN models.

%
\section{Conclusion}
\label{sec:conclusion}
In this work, we have presented a new approach for classifying RNA based on their secondary structure through image classification. By treating the BPPM representation of the secondary structure as an image, our approach takes advantage of the high speed and powerful feature extraction capabilities of CNNs. We have demonstrated this approach to be a promising way to advance RNA analysis by providing a tool for more accurate and faster RNA classification. The developed dataset can be taken as a benchmark set for any learning-based research on RNA classification. Since the images generated from RNA sequences have a special property that is significantly different from the natural images, we are currently working on developing new deep learning models that can exploit this property more effectively. 

\section*{Acknowledgement}
The work was supported in part by Nvidia GPU grant.



\begin{thebibliography}{10}

	\bibitem{$cnnclust$}
	G.~Aoki and Y.~Sakakibara, ``Convolutional neural networks for classification
	of alignments of non\-coding rna sequences,'' \emph{Bioinformatics}, vol.~34,
	no.~13, p. i237–i244, 2018.
	
	\bibitem{cen2019boosting}
	F.~Cen and G.~Wang, ``Boosting occluded image classification via subspace
	decomposition-based estimation of deep features,'' \emph{IEEE transactions on
		cybernetics}, 2019.
	
	\bibitem{$resnet$}
	K.~He, X.~Zhang, S.~Ren, and J.~Sun, ``Deep residual learning for image
	recognition,'' \emph{2016 IEEE Conference on Computer Vision and Pattern
		Recognition (CVPR)}, 2016.
	
	\bibitem{he2018spindle}
	L.~He, M.~Yu, and G.~Wang, ``Spindle-net: Cnns for monocular depth inference
	with dilation kernel method,'' in \emph{2018 24th International Conference on
		Pattern Recognition (ICPR)}.\hskip 1em plus 0.5em minus 0.4em\relax IEEE,
	2018, pp. 2504--2509.
	
	\bibitem{$graphclust$}
	S.~Heyne, F.~Costa, D.~Rose, and R.~Backofen, ``Graphclust: alignment-free
	structural clustering of local rna secondary structures,''
	\emph{Bioinformatics}, vol.~28, no.~12, p. i224–i232, Sep 2012.
	
	\bibitem{$rnafold$}
	I.~L. Hofacker, ``Vienna rna secondary structure server,'' \emph{Nucleic Acids
		Research}, vol.~31, no.~13, p. 3429–3431, Jan 2003.
	
	\bibitem{$rfam$}
	I.~Kalvari, J.~Argasinska, N.~Quinones-Olvera, E.~P. Nawrocki, E.~Rivas, S.~R.
	Eddy, A.~Bateman, R.~D. Finn, and A.~I. Petrov, ``Rfam 13.0: shifting to a
	genome-centric resource for non-coding rna families,'' \emph{Nucleic Acids
		Research}, vol.~46, no.~D1, Mar 2017.
	
	\bibitem{$graph_2004$}
	Y.~Karklin, R.~F. Meraz, and S.~R. Holbrook, ``Classification of non-coding rna
	using graph representations of secondary structure,'' \emph{Biocomputing
		2005}, 2004.
	
	\bibitem{$alexnet$}
	A.~Krizhevsky, I.~Sutskever, and G.~E. Hinton, ``Imagenet classification with
	deep convolutional neural networks,'' \emph{Communications of the ACM},
	vol.~60, no.~6, p. 84–90, 2017.
	
	\bibitem{li20202}
	K.~Li, W.~Ma, U.~Sajid, Y.~Wu, and G.~Wang, ``Object detection with
	convolutional neural networks,'' \emph{Deep Learning in Computer Vision:
		Principles and Applications}, vol.~30, no.~31, p.~41, 2020.
	
	\bibitem{ma2020mdfn}
	W.~Ma, Y.~Wu, F.~Cen, and G.~Wang, ``Mdfn: Multi-scale deep feature learning
	network for object detection,'' \emph{Pattern Recognition}, vol. 100, p.
	107149, 2020.
	
	\bibitem{$mccaskill$}
	J.~S. Mccaskill, ``The equilibrium partition function and base pair binding
	probabilities for rna secondary structure,'' \emph{Biopolymers}, vol.~29, no.
	6-7, p. 1105–1119, 1990.
	
	\bibitem{patel2020comparative}
	K.~Patel, K.~Li, K.~Tao, Q.~Wang, A.~Bansal, A.~Rastogi, and G.~Wang, ``A
	comparative study on polyp classification using convolutional neural
	networks,'' \emph{PloS one}, vol.~15, no.~7, p. e0236452, 2020.
	
	\bibitem{$convolutional_review$}
	W.~Rawat and Z.~Wang, ``Deep convolutional neural networks for image
	classification: A comprehensive review,'' \emph{Neural Computation}, vol.~29,
	no.~9, p. 2352–2449, 2017.
	
	\bibitem{$qrna$}
	E.~Rivas and S.~R. Eddy, ``Noncoding rna gene detection using comparative
	sequence analysis,'' \emph{BMC Bioinformatics 2}, vol.~8, Oct 2001.
	
	\bibitem{sajid2020zoomcount}
	U.~Sajid, H.~Sajid, H.~Wang, and G.~Wang, ``Zoomcount: A zooming mechanism for
	crowd counting in static images,'' \emph{IEEE Transactions on Circuits and
		Systems for Video Technology}, 2020.
	
	\bibitem{$sankoff$}
	D.~Sankoff, ``Simultaneous solution of the rna folding, alignment and
	protosequence problems,'' \emph{SIAM Journal on Applied Mathematics},
	vol.~45, no.~5, p. 810–825, 1985.
	
	\bibitem{$vggnet$}
	K.~Simonyan and A.~Zisserman, ``Very deep convolutional networks for
	large-scale image recognition,'' \emph{ICLR}, 2015.
	
	\bibitem{$dotaligner$}
	M.~A. Smith, S.~E. Seemann, X.~C. Quek, and J.~S. Mattick, ``Dotaligner:
	identification and clustering of rna structure motifs,'' \emph{Genome
		Biology}, vol.~18, no.~1, 2017.
	
	\bibitem{$carna$}
	D.~A. Sorescu, M.~Mohl, M.~Mann, R.~Backofen, and S.~Will, ``Carna--alignment
	of rna structure ensembles,'' \emph{Nucleic Acids Research}, vol.~40, no.~W1,
	Nov 2012.
	
	\bibitem{$googlenet$}
	C.~Szegedy, W.~Liu, Y.~Jia, P.~Sermanet, S.~Reed, D.~Anguelov, D.~Erhan,
	V.~Vanhoucke, and A.~Rabinovich, ``Going deeper with convolutions,''
	\emph{2015 IEEE Conference on Computer Vision and Pattern Recognition
		(CVPR)}, 2015.
	
	\bibitem{$rnaz$}
	S.~Washietl, ``Prediction of structural noncoding rnas with rnaz,''
	\emph{Comparative Genomics}, p. 503–526.
	
	\bibitem{$locarna$}
	S.~Will, T.~Joshi, I.~L. Hofacker, P.~F. Stadler, and R.~Backofen, ``Locarna-p:
	Accurate boundary prediction and improved detection of structural rnas,''
	\emph{Rna}, vol.~18, no.~5, p. 900–914, 2012.
	
	\bibitem{xu2019adversarially}
	W.~Xu, S.~Keshmiri, and G.~Wang, ``Adversarially approximated autoencoder for
	image generation and manipulation,'' \emph{IEEE Transactions on Multimedia},
	vol.~21, no.~9, pp. 2387--2396, 2019.
	
	\bibitem{$wrn$}
	S.~Zagoruyko and N.~Komodakis, ``Wide residual networks,'' \emph{Procedings of
		the British Machine Vision Conference 2016}, 2016.
	
	\bibitem{$review$}
	Y.~Zhang, H.~Huang, D.~Zhang, J.~Qiu, J.~Yang, K.~Wang, L.~Zhu, J.~Fan, and
	J.~Yang, ``A review on recent computational methods for predicting noncoding
	rnas,'' \emph{BioMed Research International}, vol. 2017, p. 1–14, 2017.
	
	\bibitem{$mfold$}
	M.~Zuker, ``Mfold web server for nucleic acid folding and hybridization
	prediction,'' \emph{Nucleic Acids Research}, vol.~31, no.~13, p. 3406–3415,
	Jan 2003.
	
\end{thebibliography}

\end{document}